# Predicting Social Perception from Faces – A Deep Learning Approach

Uwe Messer[1] and Stefan Fausser[2]

## Abstract
Warmth and competence represent the fundamental traits in social judgment that determine emotional reactions and behavioral intentions towards social targets. This research investigates whether an algorithm can learn visual representations of social categorization and accurately predict human perceivers' impressions of warmth and competence in face images. In addition, this research unravels which areas of a face are important for the classification of warmth and competence. We use Deep Convolutional Neural Networks to extract features from face images and the Gradient-weighted Class Activation Mapping (Grad-CAM) method to understand the importance of face regions for the classification. Given a single face image the trained algorithm could correctly predict warmth impressions with an accuracy of about 90% and competence impressions with an accuracy of about 80%. The findings have implications for the automated processing of faces and the design of artificial characters.

## Keywords
stereotypes, impression formation, face, warmth, competence, convolutional neural network, machine learning

Physiognomy, which dates back to the time of ancient Greece, is the practice of reading traits from faces. Still today, many people believe that faces reveal important traits and that by merely looking at somebody one can derive important information about the target's underlying personality (Hassin and Trope 2000). Forming impressions of other people and even anticipating how others would evaluate a person based on the person's facial appearance is relevant in a myriad of situations. For instance, recruiters have to decide whom to invite for a job interview based on curriculum vitae and a picture of the applicant. They are also confronted with facial impressions of potential employees who market themselves on social networking platforms such as Facebook and business networking platforms such as LinkedIn. Entrepreneurs provide profile photos on crowdfunding platforms to influence backers' investment decisions. Marketers carefully select facial imagery for product and brand communication and people who are seeking a partner online form impressions based on pictures of potential dating partners and decide whom they would like to meet in real life based on these impressions. Moreover, since encounters with other people today are often computer-mediated, and because these computer-mediated environments usually lack social information and put limits on social interaction (Riegelsberger and Sasse 2001), companies are eager to add social presence through facial imagery and recommendation agents in order to leverage positive experiences and increase conversion rates on commercial websites (Holzwarth et al. 2006; Kim and Park 2017). Understanding what impression a face would make on a user or customer is therefore more important than ever.

Previous work shed light on how people derive person impressions from faces and primarily focused on the question whether human perceivers can or cannot detect underlying personality characteristics from faces (Holtzman 2011; Gorn et al. 2008). Besides, research used morphing to generate prototypical faces for certain character traits (e.g., Imhoff et al. 2013). Recently, face research using machine learning methods has attracted much attention in the psychology (e.g., O'Toole et al. 2018), information systems (e.g., Kute et al. 2019) and the information sciences literatures (e.g., Antipov et al. 2017). These research streams made substantial contributions in the domains of face recognition and the prediction of human demographics from faces. However, previous work has not investigated whether an algorithm could learn human social perception and reliably predict how other people would judge personality traits of targets based on facial images alone. This would broaden our knowledge on visual representations of stereotypes in the face, clarify if human social perception can be learned by an algorithm, and would potentially improve our understanding of features in faces relevant for social categorization. The present research focuses on the training of an algorithm that predicts how faces would be perceived by human judges along the two primary dimensions of social perception: warmth and competence

[1] Department of Sales and Marketing, University of Bamberg, 96052 Bamberg, Germany
[2] Institute of Neural Information Processing, University of Ulm, 89069 Ulm, Germany

(Fiske et al. 2002). To achieve this goal, we trained two Convolutional Neural Networks (CNNs) to extract features from facial images. Afterwards, we applied the Gradient-weighted Class Activation Mapping (Grad-CAM) method (Selvaraju et al. 2017) for visualizing the regions of a face (e.g., nose, mouth, chin, etc.) that were most important for the classification of traits for the model. We present initial empirical results and lay out an agenda for further research.

The initial findings of this research have three major implications. First, we demonstrate that by using a machine vision framework, based on deep learning, it is possible to accurately predict how other people would evaluate a human face in terms of warmth and competence. Second, we find that the classification of warmth in faces is more accurate than the classification of competence. This is in line with research on human perceivers, indicating that warmth features receive more attention than competence features (Fiske et al. 2007). Third, this research helps to identify core areas of the face which are relevant for impression formation and contributes to research on explainable machine learning. Findings of this research have important practical implications. First, the trained model can help automate face processing and support decision makers choose faces that convey an intended impression on others. For example, our model could assist with employee selection, increasing efficacy in the pre-selection of applicants who would make a competent and trustworthy impression on other people. Further, by unraveling which features of a face are most important for impression formation our work could guide the design of artificial faces of avatars, computer-generated characters, as well as characters that represent brands and companies, such as Mr. Clean.

## Theoretical Foundation

### Fundamental Dimensions of Social Judgment

Our work is rooted in socio-psychological literature and the Stereotype Content Model in particular. To navigate the social world individuals need to anticipate others' intentions towards them and their capability to execute these intentions. Thus, judgments of other people are made along two primary trait dimensions: warmth and competence (Fiske et al. 2002). Warmth reflects a favorable orientation towards others and captures traits that are related to perceived intent of a person, including friendliness, helpfulness, sincerity, trustworthiness, and morality. Competence reflects traits which are related to perceived ability of a person, including intelligence, confidence, creativity, and efficacy (Fiske et al. 2007). These two dimensions "account almost entirely for how people characterize others" (Fiske et al. 2007, 77). Research on social cognition concludes that interpersonal warmth is the primary trait. Information about other's intentions (is the person well intended or ill-intended) is more important for deciding about approach and avoidance than information about others' capabilities. Therefore, warmth judgments are dominant and receive more weight in interpersonal perception than competence judgments (Fiske et al. 2007).

The perceptions of warmth and competence originate from socio structural variables. On a broader level person impressions are driven by perceptions of competition and status. Competition refers to incompatibility of a person's goals to the goals of the perceiver. Status is reflected by prestige and societal resources of an individual (Fiske et al. 2002; Fiske et al. 2007; Fiske and Ruscher 1993). Competition is leveraged when others could potentially harm the judges' goals (Fiske et al. 2007), and individuals with competitive intentions (i.e., the intent to interfere with goals) are perceived as cold, whereas individuals with cooperative intentions are perceived as warm. People having the ability to achieve their intentions and assert their goals are seen as competent, whereas those who are not are seen as being incompetent (Fiske et al. 2002).

Turning to the outcomes, distinct emotions and behavioral intentions follow from the four possible competence-by-warmth combinations (Fiske et al. 2002). Warmth leads to active facilitation. Individuals who are perceived as warm are more likely to receive help from others, while those seen as cold are more likely to receive negative responses. Competent individuals are often chosen as cooperation partners while people who seem to lack competence receive passive harm, such as ignoring. While individuals perceived as warm and competent elicit uniformly positive emotions, such as admiration, people perceived as lacking warmth and competence elicit uniformly negative emotions, such as contempt. Neuroscience studies show that people who are perceived as cold and incompetent get dehumanized (Harris and Fiske 2006), whereas those who are perceived as either warm but not competent or cold but competent, so-called mixed judgments, cause ambivalent emotional reactions (pity and envy; Fiske et al. 2007). In sum, research demonstrates that warmth and competence are the key traits predicting emotional and behavioral intentions towards other people and are therefore relevant in a wide range of settings, including face-to-face and computer-mediated interactions.

**Impression Formation from Faces**

Faces provide a focal point of interaction in human-to-human encounters and get handled by discrete neural areas in the brain (Kanwisher and Yovel 2006). Considerable evidence suggests that people are using facial properties to make judgments about characteristics of social targets (Engell et al. 2007) and that these judgments are largely automatic and unintentional (Farah 1996). For instance, people with more symmetrical faces are perceived as less neurotic, more agreeable, and more conscientious than their counterparts with less symmetrical faces (Noor and Evans 2003). Symmetry is one of the cues for facial attractiveness and attractive people are inferred to possess more positive character traits such as sociability and inherent goodness (Langlois et al. 2000).

Some facial features are sexual dimorphic meaning that they are expressed more strongly in one gender then the other. For example, men with wider faces (i.e., who have a higher ratio of the distance between the left and right cheekbones to the distance between mid-brow and upper lip) are perceived as more threatening (Geniole et al. 2015). Moreover, people entrust more money to other players with narrower faces compared to those with wider faces because these players are seen as less trustworthy (Stirrat and Perrett 2010). Wider faces make a target look more robust and imposing and thus reduces the danger of physical confrontation, incentivizing self-interested behavioral tendencies of men with wider faces (Stirrat and Perrett 2010). Learning processes might then have led to stereotypes about these men. Another cue that people use to make judgments about others is "babyfacedness". A baby-face is characterized by features commonly found in faces of babies and young children, such as large eyes, small chin, and tiny nose (Gorn et al. 2008). Baby-faced individuals are perceived as more honest, sincere, and more trustworthy than their mature-faced counterparts (Gorn et al. 2008). Structural features of the face are also used to infer so-called dark personality traits such as narcissism and psychopathy (Holtzman 2011). In one study, observers who saw emotionally-neutral faces of men and women made spontaneous conclusions about underlying dark traits of the targets. Sharp facial lines such as clear angles were used as indicators for the presence of these traits (Holtzman 2011). Research also found that people use structural differences in faces to judge aggressive nature of a target. Horizontally narrowed eyes and lower levels of fatty deposits in the chin and cheekbone areas were used as indicators of trait aggression (Trebický et al. 2013).

Besides these rather morphological features, emotional expressions also convey information about underlying traits of the expresser. Emotions have evolved to assist social interaction and human perceivers are skilled interpreters of emotions in the face (Keltner and Haidt 1999). Smiles for example, are signaling cooperative and friendly intentions and thus convey interpersonal warmth (Kraus and Chen 2013). A broader smile (i.e., a higher intensity smile) signals more friendliness and sociability, but also more submissiveness, thus rendering the expresser less competent (Wang et al. 2016). Emotions are often overgeneralized and used to form first impressions. For instance, there is a tendency to infer that a sad looking person has a submissive character (Montepare and Dobish 2003). Taken together, prior research indicates that people "read" faces and use morphological features and emotional expressions to form impressions about targets' personality traits including impressions of warmth (e.g., face width, facial maturity, sharp facial lines) and competence (e.g., smile intensity, eye shape).

## The Present Research

Prior research concludes that warmth and competence are the primary dimensions of social judgment and that people use facial features to form impressions about these two dimensions. In the next sections, we explore how well warmth and competence can be automatically detected in face images using machine learning. As research on impression formation and social perception postulate that warmth receives more weight than competence in human encounters, we investigate whether warmth and competence can be equally well detected using machine learning or not. These two questions refer to classification performance of the models. Finally, we try to uncover the areas of faces which are more or less important for the categorization of the outcomes.

## Methodology

**Dataset and Measures of the Outcomes**

We used the 10K Adults Face database (Bainbridge et al. 2013). The database contains more than 10,000 face images in JPEG format with 96 pixels per inch resolution and 256-pixel height. 2,222 of these images were evaluated by 15 human raters each. Amongst other psychological measures, raters evaluated how warm and how confident they perceived the targets on the images. Raters were asked: "Do you think the face looks cold?" and "Do you think the face looks confident?" and provided their answer on a 9-point Likert-type rating scale ranging from 1 = "*not at all*" to 9 = "*extremely*." We decided to use these two variables, cold and confident, as proxy variables for the two fundamental dimensions of social perception. As an initial step, we decided to dichotomize the outcome variables in order to achieve more accurate classifications. Therefore, we used the 10% and the 90%-percentile of each of the two outcome variables and retained only those face images which lied within these percentiles resulting in a final set of 478 (warmth, inverse of the cold variable) and 482 (competence, with confidence as proxy) images. Out of the 478 warmth images, 222 are in the 90% percentile (cold) and 256 are in the 10% percentile (warm). Out of the 482 competence images, 256 are in the 90% percentile (competent) and 226 are in the 10% percentile (incompetent). Further, each of the two datasets (warmth and competence) were randomly split into training and validation sets with equal probability.

| Demographics | Warmth | | Competence | |
|---|---|---|---|---|
| | Training sample | Validation sample | Training sample | Validation sample |
| Number of images | 239 | 239 | 241 | 241 |
| Median age | 30 to 45 years | 30 to 45 years | 30 to 45 years | 30 to 45 years |
| Female | 44% | 44% | 39% | 42% |
| Race | | | | |
|   White | 82% | 74% | 79% | 81% |
|   Black | 10% | 19% | 11% | 13% |
|   Asian | 4% | 2% | 2% | 2% |
|   Hispanic | 4% | 4% | 5% | 3% |
|   Other | 0% | 1% | 3% | 1% |
| Notes: Demographics of individuals shown on images were determined by raters. | | | | |

**Table 1. Sample Demographics of the Face Images**

**Classification Using Convolutional Neural Networks**

Face images were classified using Convolutional Neural Networks (CNNs). We briefly introduce CNN principles before we describe the architecture of the CNNs used. In deep learning, multiple processing layers are employed to learn representations of data and to combine these representations in order to classify the data. CNNs represent a class of feedforward deep neural network (DNN), where the first layers uses convolutional filters, also known as kernels, to extract important features from previous layers and create increasingly abstract representations of the input stimulus. The last layers are fully-connected layers that weigh and combine the former representations and eventually end up in an output that allows for data classification. By minimizing a loss function, the convolutional filters and the neuron weights in the fully-connected layers are learned automatically, thus, being more efficient than handcrafting the features. CNNs are a state-of-the art method for image recognition and classification (Goodfellow et al. 2016) and represent one of the most powerful deep learning approaches (Silver et al. 2016). Due to their complexity, CNNs are black-box methods for human perceivers because they cannot understand what features are used to make a prediction or other decisions in the network. Recently, the Gradient-weighted Class Activation Mapping (Grad-CAM) method (Selvaraju et al. 2017) was introduced which aims to produce visual explanations for the decisions of the CNNs. The method uses the class-specific gradients of the last convolutional layer for localizing the important regions in the image.

CNNs differ by architecture, which refers to the number, type, order, and size of their layers. The architecture of our CNN is as follows: We use three convolutional layers with Rectified Linear Units (ReLU) activations where the first two have 32 convolutional filters of size 3x3 each and the last has 64 filters of size 3x3. After the ReLU function, each convolution layer is max pooled with a pool size of 2x2. Then we have one fully-connected layer, ReLU activation and 64 neurons, and one fully-connected layer, softmax activation and 2 neurons. Each of the two neurons in the final layer have an output between 0 and 1, and the sum of the two outputting neurons must be 1. Accordingly, these outputs are interpreted as the estimated probability of an image belonging to a certain class (e.g., warm or cold). We used two separate CNNs with binary classification of the outcomes (warmth and confidence). The model was implemented in Python and the code can be requested from the authors.

Because of the limited number of training images, the training images were randomly pre-processed including brightness modifications (80% to 120%), clockwise and anti-clockwise rotations (up to 20%) and zoom-in and zoom-out (up to 20%) operations. With these random preprocessing operations, the CNN learned to handle variations, i.e. kept more generic. Please note that the validation images were kept untouched.

## Results

Our models took an image as input and outputted two probabilities reflecting whether the face displayed on the image was warm vs cold or competent vs incompetent, respectively. Regarding classification performance, both CNNs had high accuracy. The CNN for classifying warmth achieved a classification accuracy of >90%. The CNN for classifying competence achieved a classification accuracy of >80%. The classification of warmth was superior to the classification of competence. This was expected as the human perceivers focus more on the warmth feature than the competence feature (Fiske et al. 2007).

In the next step, we inspected the heat maps resulting from the Grad-CAM algorithm. One example can be found below (see Figure 1). The heat maps highlight regions of the face images that were more important (red color) or less important (blue color) for the classification of the outcomes. Heat maps reveal that for warmth, classification mostly relied on eyes and mouth, especially the lower lip. In some instances, regions in the infraorbital triangle (i.e., the region below the eye socket) mostly close to the nostril wings such as the nasolabial fold were highlighted. Rarely, classification also relied on the right or the left facial boundary. For competence, heat maps revealed that the model relied mostly on the area around the eyes such as the lower eyelid and the skin fold in the upper eyelid. Often, the heat map also highlighted the area between the upper lip and the nose (i.e., the philtrum) and the nostrils.

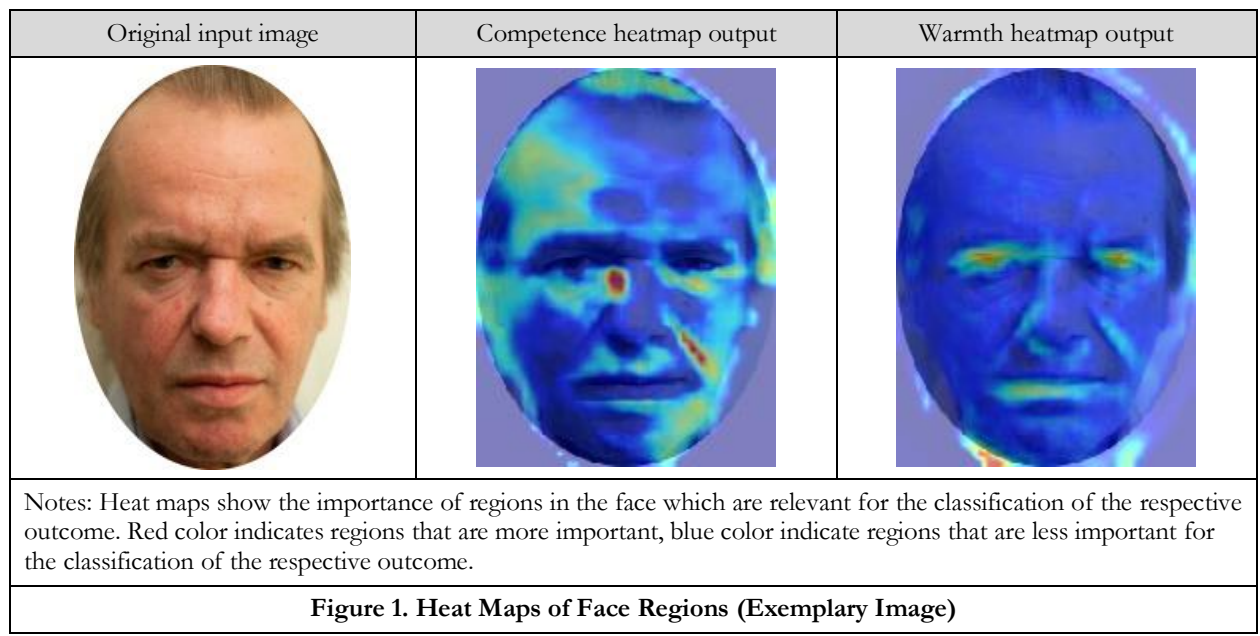

| Original input image | Competence heatmap output | Warmth heatmap output |

Notes: Heat maps show the importance of regions in the face which are relevant for the classification of the respective outcome. Red color indicates regions that are more important, blue color indicate regions that are less important for the classification of the respective outcome.

**Figure 1. Heat Maps of Face Regions (Exemplary Image)**

## Discussion

### Contributions of the Research

This research demonstrates that a machine can learn visual representations of social categories and accurately predict human perceivers' impressions of face images. Our initial results show that an algorithm can accurately classify warmth and competence in faces and that warmth can be classified with higher accuracy than competence. Finally, we use heat maps to highlight regions in the face that are most indicative for the classification of warmth and competence, respectively.

The current research adds to the work on visual representations of stereotypes in the human face (e.g., Dotsch et al. 2013; Imhoff et al. 2013). Prior studies used image classification and face morphing to understand how people visually encode impressions. Face morphs, which represent prototypical faces, have limited value for the localization of specific facial features relevant for impression formation. Using a machine learning approach we go one step further and pinpoint those facial features that were most relevant for the classification task. Thereby, we uncover subtle, yet unknown differences used to infer warmth vs competence. For instance, we find that the pupils bear more weight for the classification of warmth than the area around the eyes while the latter area is more important for the classification of competence than the pupils. Further, we found a link between emotional expressions used to communicate confrontational tendencies, such as anger and contempt, and impression formation. In many faces that were accurately classified as cold the algorithm relied on those facial areas which are associated with contempt and anger, such as lips, nostrils, and mouth corners (see Ekman and Friesen 2003). Further, this research adds to the work on explainable machine learning by visualizing important classification features in our deep neural networks via heat maps (e.g., Setiono et al. 2000).

This research also has important practical implications. First, since task automation due to technological innovation is a key driver of the economic system, our model can leverage efficiency by taking over face classification tasks. For instance, the model could be employed to automate the processing of job applications, assisting experts by pre-selecting those applicants who have the highest possibility of making favorable impressions on others. This kind of pre-selection is most relevant in areas were first impressions are crucial for success, such as in sales or service. Second, using the algorithm to identify people of varying combinations of warmth and competence could be useful in other areas as well. For instance, since warmth impressions predict behavior towards the target, charities and disaster relief agencies could employ our model to select stimuli of people who would receive active facilitation. Pictures of these people might be more effective in a campaign that aims towards the elicitation of empathy and donations. Likewise, this approach could also guide the selection of avatars or facial representations of chat bots that assist users on websites. Previous research has shown that such avatars and chat bots are often selected in a way to maximize the impression of warmth and/or competence (Woelfl and Feste 2018). Third, the model presented in this research is also useful in the domain of stigmatization and could assist with the detection of individuals who would likely be victimized or might suffer from neglect or social exclusion (i.e., individuals low on the warmth dimension). Earlier detection of individuals at risk might help to leverage preventive measures. Finally, by pinpointing face features indicative of warmth and competence this research helps to deliberately design artificial human-like characters such as brand and company spokespersons (e.g., Mr. Clean), artificial characters in movies and video games, or even humanoid robots endowed with human-like faces.

**Avenues for Further Research**

Although our research has offered novel insight into the social categorization of faces, more research is still needed. We identified five promising avenues for further research. First, we used a categorical outcome treating warmth and competence as either being high or low. In a next step, a gradual prediction of the outcomes would be promising. By doing so, we could better understand which facial features are used to distinguish a slightly warm/competent looking face from a very warm/competent looking face. Probably different face regions might be of different importance for the different warmth/competence degrees. Second, additional studies might want to test the empirical reproducibility of the findings reported. To this end, future research endeavors could use a larger sample of input faces and could test whether a machine could also predict social impressions from emotionally neutral faces (i.e., faces without visible emotions). It is possible that the categorization by both, human raters and the trained algorithm, is biased by emotional expressions in the faces in the current sample. Third, the current research used social impression ratings by human perceivers as outcome labels. Prior research concluded that rater personality biases the interpretation of faces and raters could, for instance, vary regarding their perceptions of anger (e.g., Maner et al. 2005). Therefore, future research could use a larger number of raters for each input face. This would allow for more generalizable conclusions since more raters could eliminate biases commonly found in small samples. Fourth, a multi-method approach is suggested. For instance, the validity of the facial features derived from the heat maps could be tested using experimental manipulations of faces. Further research might want to purposefully manipulate these facial features in existing face images or generate artificial face images which are then evaluated by human raters to test their validity. Fifth and finally, since we focused our research on human faces, future studies might want to extend our methodology to investigate the social perception of non-human faces as found in artificial characters (Messer et al. 2009a, Messer et al. 2009b). This extension seems natural given that neuroscientific evidence shows that making judgements about artificial faces activates the same neural systems that are activated when judging human faces (Gazzola et al. 2007).